%% file: acl_latex.tex
\documentclass[11pt]{article}

\PassOptionsToPackage{table}{xcolor}
\usepackage[final]{acl}

\usepackage{times}
\usepackage{latexsym}

\usepackage[T1]{fontenc}

\usepackage[utf8]{inputenc}

\usepackage{microtype}

\usepackage{inconsolata}

\usepackage{graphicx}

\usepackage{amsmath}
\usepackage{multirow}
\usepackage{booktabs}
\usepackage{amssymb}

\usepackage{etoolbox}
\AtBeginEnvironment{equation}{\small}
\AtBeginEnvironment{align}{\small}
\AtBeginEnvironment{gather}{\small}
\AtBeginEnvironment{multline}{\small}

\usepackage[most]{tcolorbox}
\definecolor{ourscolor}{HTML}{EDF2F7}
\title{Efficient Language-to-Vision Feature Injection for Referring\\ Single-Object Tracking}

\author{
 \textbf{Han Wang\textsuperscript{1,2}},
 \textbf{Yuxuan Liu\textsuperscript{1,2}},
 \textbf{Yuhan Sun\textsuperscript{1,2}},
 \textbf{Jian Yang\textsuperscript{1,2}},
 \textbf{Xiaotong Xu\textsuperscript{1,2}},
\\
 \textbf{Yixuan Lv\textsuperscript{1}},
 \textbf{Zhuang Zhou\textsuperscript{1}},
 \textbf{Shengyang Li\textsuperscript{1,2}}\thanks{Corresponding author}
\\
\\
 \textsuperscript{1}Technology and Engineering Center for Space Utilization, Chinese Academy of Sciences,
\\
 \textsuperscript{2}University of Chinese Academy of Sciences
}

\begin{document}
\maketitle

\input{sec/1_abstract}
\input{sec/2_introduction}
\input{sec/3_related}
\input{sec/4_method}
\input{sec/5_experiments}
\input{sec/6_conclusion}

\section*{Limitations}
Although LVTrack provides two input-resolution variants, the current implementation still relies on fixed input resolutions and has not fully explored the potential of the architecture for arbitrary-resolution inputs. In addition, this work mainly studies language guidance provided at initialization, while practical scenarios may involve dynamically updated user instructions or additional constraints during tracking. Future work could investigate arbitrary-resolution tracking and interactive language guidance to further broaden the applicability of LVTrack.

\section*{Acknowledgements}
This work was financially supported by the Science and Technology Support Special Project 3 (Project No. Y4030221WY). The authors would like to express sincere gratitude for the funding support.


\bibliography{reference}

\input{sec/7_appendix}
\end{document}

%% file: sec/1_abstract.tex
\begin{abstract}
Referring single-object tracking enables language-grounded target initialization and subsequent tracking by jointly leveraging semantic cues and visual templates.
The core difficulty is to use language differently across stages: it is indispensable for grounding but can induce semantic drift during tracking when overemphasized.
Meanwhile, current methods often require costly vision-language alignment training.
We present LVTrack, a pure transformer framework that introduces a mode-conditioned Gated Feature Injector to adaptively regulate textual guidance and alleviate semantic drift.
Together with targeted adaptations, it directly harnesses a frozen vision-language pretrained model, greatly reducing training cost and preserving strong language understanding.
To further improve temporal localization, LVTrack integrates hybrid relative-absolute positional encodings with a lightweight memory mechanism and optimizes autoregressive box prediction using a Gaussian-smoothed KL loss.
Extensive experiments on standard benchmarks demonstrate that LVTrack achieves strong performance.
\end{abstract}

%% file: sec/2_introduction.tex
\section{Introduction}

Recent single-object tracking (SOT) methods have increasingly explored natural-language (NL) referring expressions as a flexible interface for target specification \cite{ref1, ref2, ref3}.
Compared with conventional SOT \cite{ref4, ref5}, referring SOT (RSOT) enables more intuitive human interaction and improves robustness in challenging scenarios by leveraging semantic cues beyond visual template.
\begin{figure}[t]
\centering
\includegraphics[width=2.8in]{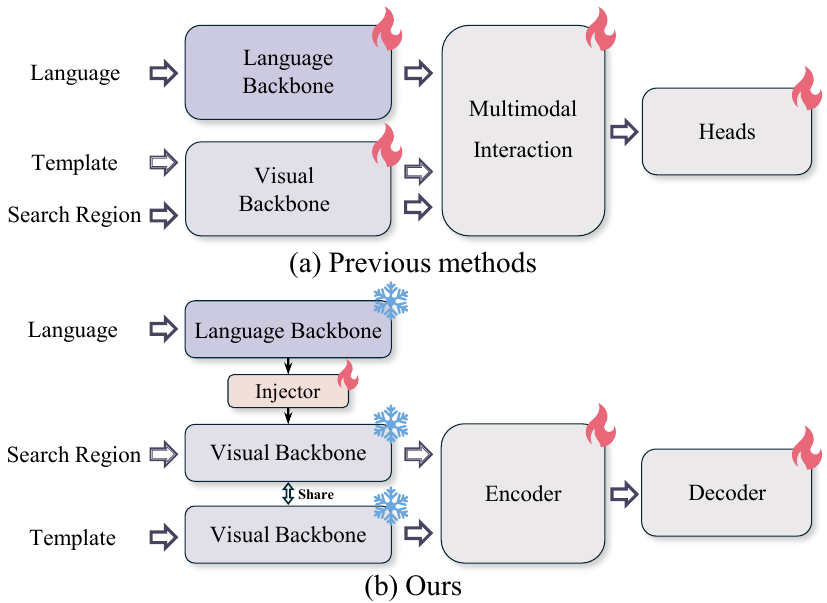}
\hfil
\vspace{-1ex}
\caption{Architectural comparison between LVTrack and prior frameworks. LVTrack performs discrete-token autoregressive box prediction and relies on a fully frozen VLP backbone, injecting textual cues into visual features via a lightweight injector. Leveraging the pretrained backbone obviates costly vision–language alignment training, substantially reducing training cost.}
\label{pipline}
\vspace{-2.5ex}
\end{figure}

RSOT is commonly decoupled into two sub-tasks: (i) visual grounding and (ii) visual tracking.
Recent works have sought to strengthen the coupling between grounding and tracking through end-to-end multimodal reasoning, typically initializing the language backbone with BERT \cite{ref8} alongside a dedicated visual feature extractor.
By jointly modeling and aligning visual and textual representations during training, a single model can localize the object from either a language description or an image template.
However, robust vision–language alignment generally hinges on large-scale image–text contrastive data and long-horizon pretraining \cite{ref9}, whereas existing approaches are trained primarily on tracking and grounding datasets whose scale is insufficient to support learning alignment at a general-purpose level.

Large-scale vision–language pretraining (VLP) models (e.g., CLIP \cite{ref10}, SigLIP \cite{ref11}) acquire strong cross-modal alignment and generalizable priors via image–sentence contrastive learning.
However, their alignment lacks explicit instance-level binding and spatial selection within an image, and thus cannot directly support object-level discrimination and localization under referring expressions.
Consequently, directly feeding visual and language backbone features into a learnable multimodal interaction encoder typically still requires end-to-end optimization of the backbone or a high-capacity fusion module \cite{ref46, ref45}, incurring substantial training cost and exacerbating overfitting on limited tracking data.
To address this, we freeze the VLP backbone and introduce a lightweight feature injector that performs conditional fusion via cross-attention (Fig. \ref{pipline}), selectively aggregating language evidence at each spatial position based on similarity to produce language-conditioned visual representations and strengthen target cues.
This design achieves instance-level alignment with few trainable parameters, reducing training cost without sacrificing the generalization benefits of VLP.

Another major challenge lies in balancing linguistic guidance and visual references.
The initial language description specifies the target, but may not provide reliable cues for subsequent frame-by-frame tracking.
Thus, over-reliance on linguistic priors can induce semantic drift toward semantically similar distractors \cite{ref12}, whereas excessive suppression of language cues forfeits their complementary robustness benefits.
To address this issue, we introduce a Gated Feature Injector.
The injector employs amplitude-calibrated, mode-conditioned gating, enforcing conservative textual injection for tracking while allowing stronger language-vision coupling for grounding.
A magnitude-aware regularizer adaptively calibrates the gating parameters by constraining the relative gating-response norm across modes.
This asymmetric fusion suppresses ambiguity-induced linguistic interference without degrading grounding capability, improving temporal stability.

Although discretizing coordinates into categorical tokens enables autoregressive visual localization \cite{ref6, ref14}, standard cross-entropy is suboptimal because it ignores the spatial topology among bins and penalizes near and distant errors equally.
To address this, we introduce a Gaussian-Smoothed KL loss (GS-KL), which replaces one-hot labels with distance-aware Gaussian soft targets.
By injecting spatial priors into token classification, GS-KL reduces quantization ambiguity and equips autoregressive prediction with regression-like distance sensitivity, enabling more robust and fine-grained localization.

To improve geometric reasoning and localization accuracy, we adopt a seq2seq-style \cite{ref44} encoder-decoder architecture with hybrid positional modeling, combining learnable 2D absolute embeddings with 2D RoPE-based relative geometry inside attention \cite{ref15}.
We further introduce a lightweight memory module that carries object-specific appearance cues across frames by extracting fixed-length tokens from the previous target region as temporal priors for next-frame localization.

Motivated by these observations, we present a unified RSOT framework.
Compared with prior methods, it requires substantially lower training cost while delivering strong performance across visual grounding, NL-initialized SOT, and NL+BBOX-initialized SOT.
Our main contributions are as follows:
\begin{itemize}
    \item We propose LVTrack, a new RSOT framework built on a frozen vision–language pretrained model, which significantly reduces training cost and improves generalization.\vspace{-1ex}
    \item We introduce an amplitude-regularized, dual-mode Gated Feature Injector that calibrates textual injection to suppress semantic drift while preserving grounding capability.\vspace{-1ex}
    \item An ordinal-aware GS-KL loss is introduced for autoregressive coordinate-token prediction, with hybrid positional encoding and appearance memory to enhance tracking.\vspace{-1ex}
    \item Extensive experiments on vision-language tracking and visual grounding benchmarks demonstrate the competitiveness of LVTrack.
\end{itemize}

%% file: sec/3_related.tex
\section{Related Work}

\begin{figure*}[t]
\centering
\includegraphics[width=6in]{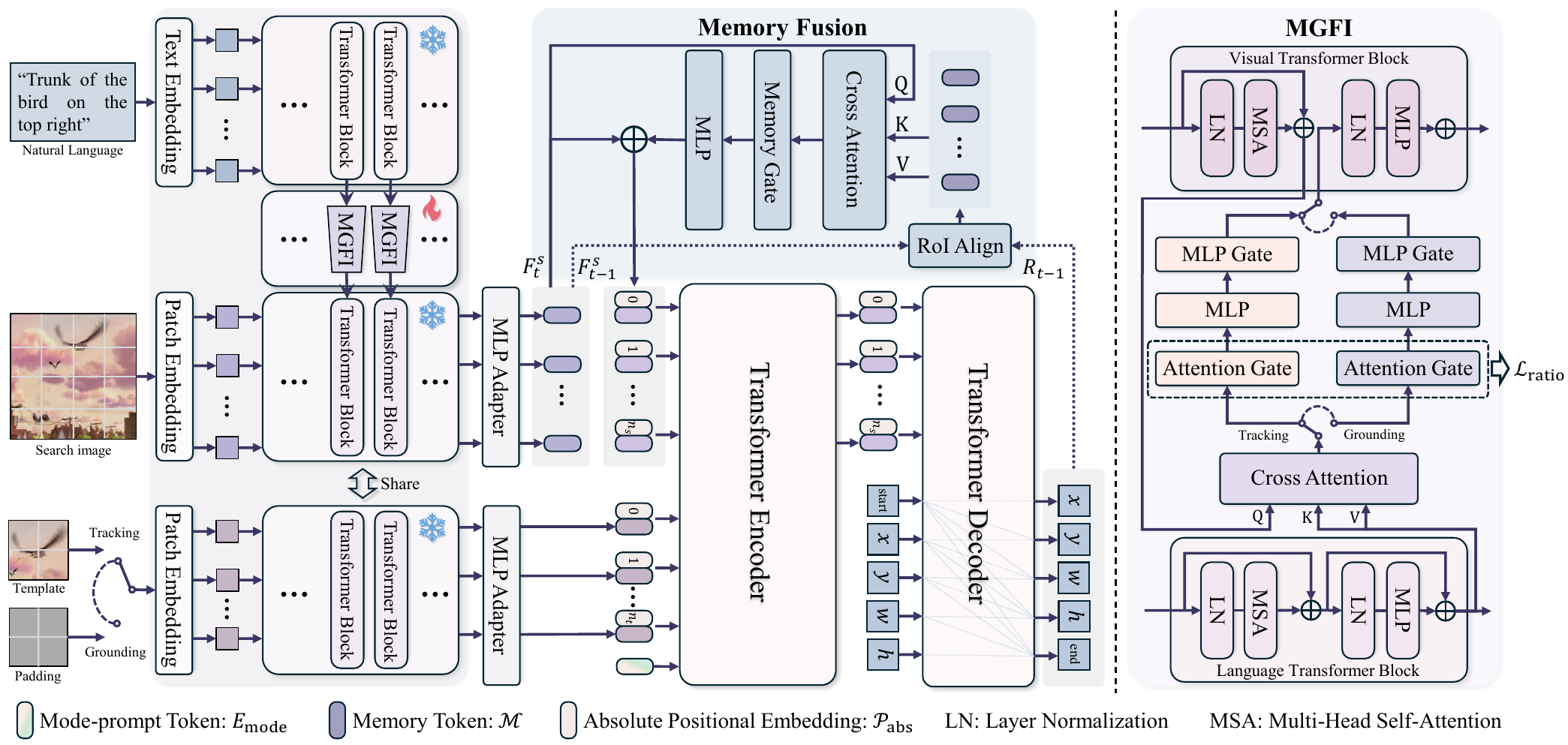}
\hfil
\vspace{-1.4ex}
\caption{Framework of proposed LVTrack. LVTrack is an end-to-end vision-language tracker that leverages a frozen VLP backbone and MGFI to enable multimodal reasoning. A position-enhanced encoder models template and search tokens, while a masked decoder autoregressively predicts discretized tokens for the object box's relative coordinates. (Residual connections in MGFI are omitted for clarity.)}
\vspace{-2.2ex}
\label{Network}
\end{figure*}

\subsection{Language-assisted SOT}

In language-assisted tracking, the object is typically specified in the first frame by a BBOX together with an accompanying language description.
In this setting, language is often treated as an auxiliary cue that complements the visual template, extending conventional BBOX-only tracking to a multimodal formulation to improve object understanding and discriminability.
SNLT \cite{ref16} introduces language guidance into a Siamese tracking framework.
VLT \cite{ref17} learns adaptive vision-language representations via a ModaMixer module and an asymmetric search strategy.
DUTrack \cite{ref3} leverages large language models to generate dynamic descriptions and further incorporates a dynamic update mechanism.
SeqTrackv2 \cite{ref19} and SUTrack \cite{ref20} concatenate NL and visual features and feed them into a unified encoder to enable cross-modal interaction, while also supporting additional modalities.
Overall, incorporating NL cues generally improves performance by enhancing the model's semantic awareness of the object.

\subsection{Language-initialized SOT}

Language-initialized SOT was first formulated and empirically validated in 2017 \cite{ref21}.
Early approaches typically adopted a two-stage pipeline, where visual grounding is performed to obtain an initial target localization, followed by conventional tracking in subsequent frames \cite{ref22}.
TNL2K \cite{ref23} not only introduced a new benchmark, but also proposed an adaptive switching framework that alternates between language-driven global search and vision-based local matching.
JointNLT \cite{ref1} took a step toward unification by presenting a joint framework with end-to-end modeling and training.
QueryNLT \cite{ref12} further exploits historical visual evidence to suppress language descriptions that become misaligned over time, producing more precise cues for tracking.
UVLTrack \cite{ref2} improves performance through enhanced multimodal alignment and a modality-adaptive bounding-box head.
MambaVLT \cite{ref27} explores a Mamba-based VL tracker, leveraging state-space dynamics for spatiotemporal evolution.
Despite recent progress, current approaches still underexploit the generalization potential of VLP models, often requiring costly training to establish effective vision-language alignment.

%% file: sec/4_method.tex
\section{Method}

\subsection{Overall Framework}

As illustrated in Fig. \ref{Network}, LVTrack comprises three core components: (i) a multimodal backbone equipped with a feature injector, (ii) a spatiotemporally enhanced transformer encoder, and (iii) an autoregressive localization decoder.
Given a template image $I^{t}$, a search image $I^{s}$, and a referring expression $L$, LVTrack aims to predict a normalized bounding box $B\in[0,1]^4$ for each search frame.
We first extract textual, template and search features using a VLP model:

\begin{equation}
F^L=\phi_l(L),\quad F^{t}=\phi_v(I^{t}),\quad F^{s}=\phi_v(I^{s}) ,
\end{equation}
where $\phi_l$ and $\phi_v$ denote the language and visual backbones. We adopt Perception Encoder (PE) \cite{ref28} as the frozen backbone, a CLIP-style model with stronger performance.
Because intermediate ViT layers preserve richer spatial cues \cite{ref36}, we use only the first 10 of the 12 layers, which yields better performance (see App. \ref{app:backbone} for details on the ViT layer selection).

LVTrack supports three operational modes: template-only tracking \(\mathcal{S}\), grounding \(\mathcal{G}\), and joint tracking \(\mathcal{T}\). These modes correspond to template-based tracking, language-based visual grounding, and joint language-template tracking, respectively.
The textual feature is injected into the search feature $F^{s}$ via the Mode-Conditioned Gated Feature Injector (MGFI).
Next, the template features and the text-injected search features are each passed through a Multilayer Perceptron (MLP) adapter to obtain $\tilde{F}^{t}\in\mathbb{R}^{N_t\times D}$ and $\tilde{F}^{s}\in\mathbb{R}^{N_s\times D}$, where $N_t$ and $N_s$ are the template and search token counts, and $D$ is the feature dimension.
Then, $\tilde{F}^{t}$ and $\tilde{F}^{s}$ are fed into an 8-layer transformer encoder for feature interaction, and a 4-layer transformer decoder autoregressively predicts the normalized object coordinates from the search tokens.

\subsection{Mode-Conditioned Gated Feature Injector}

Unlike prior approaches that perform multimodal interaction by concatenating visual and language tokens in a joint encoder, our method injects textual cues into visual representations via cross-attention modules inserted into selected ViT layers, conditioned on the task mode $m \in \{\mathcal{G}, \mathcal{T}\}$.
By freezing the VLP backbone and training only lightweight injectors, this design yields instance-grounded visual features while preserving pretrained generalization.
However, indiscriminate injection of textual cues $F^L$ into the visual stream risks semantic drift, where the tracker becomes susceptible to semantically similar distractors.
To reconcile the tension between the visual dominance required for tracking and the linguistic dependence of grounding, we introduce the MGFI.

\textbf{Gated injection mechanism.}
The architecture of MGFI is illustrated on the right side of Fig.~\ref{Network}.
Let $f^s_n$ denote the intermediate feature of the search image $I^s$ at visual layer $n$, and let $\tau(n)$ be the mapping function that retrieves the corresponding textual layer index (since the visual and language backbones have different depths, $\tau$ aligns layers by their relative depth).
MGFI aggregates linguistic semantics from $f^L_{\tau(n)}$ via Multi-Head Cross-Attention (MHCA) and injects them into the visual stream through a learnable channel-wise gate $\lambda$.
We equivalently rewrite the update as a learnable perturbation applied to the frozen visual backbone:

\begin{align}
\hat{f}^s_n 
&= f^s_n + \lambda_{\text{attn}}^m \odot \mathrm{MHCA}_n\!\left(f^s_n, f^L_{\tau(n)}, f^L_{\tau(n)}\right), \\
\Delta_n^m
&= \left(\hat{f}^s_n - f^s_n\right) + \lambda_{\text{mlp}}^m \odot \mathrm{MLP}_n(\hat{f}^s_n), \\
\tilde{f}^s_n
&= f^s_n + \Delta_n^m,
\end{align}
where $\odot$ denotes the channel-wise Hadamard product, $\lambda_{(\cdot)}=\tanh(\gamma_{(\cdot)})$ controls the injection amplitude, $\tilde{f}^s_n$ denotes the language-enhanced visual feature at layer $n$, and we initialize $\gamma=\mathbf{0}\in\mathbb{R}^{D}$.

This perturbation view makes the drift mechanism explicit: language injection modifies the visual feature consumed by the downstream predictor.
The gate norms therefore control the maximum language-induced perturbation in tracking, mitigating drift toward semantically similar distractors. We provide the detailed bound in App. \ref{app:mgfi_analysis}.

\textbf{Dual-mode conditioning and regularization.}
A key challenge in unified RSOT is to calibrate textual conditioning under task-dependent reliance on language.
For grounding, the expression serves as the primary localization cue.
For tracking, it provides complementary semantic guidance, but predictions must remain anchored to instance-level visual evidence from the template and temporal context.
MGFI addresses this challenge through a shared-yet-decoupled fusion design: the MHCA module is reused across modes to promote a consistent cross-modal fusion geometry and improve efficiency, while mode-specific gates $\gamma^{m}$ and MLP branches control the strength and form of textual injection.
This design preserves strong language-vision coupling for grounding and applies calibrated, conservative injection for tracking.

Since residual magnitudes vary across layers and samples, a relative gate constraint is more stable than an absolute threshold.
We introduce an \emph{Amplitude Ratio Loss} $\mathcal{L}_{\text{ratio}}$:

\begin{equation}
\mathcal{L}_{\text{ratio}}
=
\max\!\left(0,\;
\log\|\lambda^{\mathcal{T}}_{\text{attn}}\|-
\log\|\lambda^{\mathcal{G}}_{\text{attn}}\|-\log\rho\right),
\end{equation}
where $\rho \in (0, 1)$ is a decay factor, and a small constant \(\epsilon=10^{-8}\) is added inside each logarithm in implementation to prevent \(\log 0\) at initialization.
This enforces $\|\lambda^{\mathcal{T}}_{\mathrm{attn}}\|\le\rho\,\|\lambda^{\mathcal{G}}_{\mathrm{attn}}\|$, formalizing a vision-anchored tracking regime in which language acts as a calibrated semantic cue rather than a dominant signal.
This yields a principled fusion mechanism that resists semantic drift in tracking while preserving language-sensitive grounding.

\subsection{Spatiotemporal Tracking Modeling}

\textbf{Memory fusion.}
The transformer encoder integrates cross-frame appearance consistency with geometric alignment.
Given the MGFI-enhanced search feature $\tilde{F}^{s}$, temporal priors are first injected via a dynamic appearance memory $\mathcal{M}$ built by RoI Align over the predicted box on historical feature maps, yielding memory tokens $\mathcal{M}\in\mathbb{R}^{K\times D}$.
Memory fusion is implemented by cross-attention:

\begin{align}
\hat{F}^{sm} &= \lambda_{\text{mem}} \odot \mathrm{MHCA}\!\big(\tilde{F}^{s},\,\mathcal{M},\,\mathcal{M}\big), \\
\tilde{F}^{sm} &= \tilde{F}^{s} + \mathrm{MLP}(\hat{F}^{sm}),
\end{align}
where $\lambda_{\text{mem}}=\tanh(\gamma_{\text{mem}})$ is a learnable channel-wise gate, initialized with $\gamma_{\text{mem}}=\mathbf{0}$ at start.
In addition, $\mathcal{M}$ is updated at inference only when the prediction confidence exceeds a threshold $\tau_{\text{mem}}$ to avoid noisy accumulation.

\textbf{Encoder-decoder architecture.}
The encoder input is defined as

\begin{equation}
S_{\mathrm{in}}=[E_{\mathrm{mode}}(m),\,\tilde{F}^{t},\,\tilde{F}^{sm}],
\end{equation}
where $E_{\mathrm{mode}}\in\mathbb{R}^{3\times D}$ parameterizes three learnable mode prompts.
To preserve spatial coherence between template and search tokens, we introduce a shared hybrid positional scheme that combines a learnable 2D absolute coordinate anchor with 2D Axial RoPE.
The shared absolute embedding aligns template and search features under a common coordinate system, while 2D RoPE injects relative geometric bias and remains robust to mismatched token-grid scales.
Finally, a SeqTrack-style autoregressive decoder \cite{ref6}, also equipped with RoPE, predicts discretized coordinate tokens from the encoded search features.
This design couples mode-aware feature interaction with geometry-consistent localization; further positional modeling details are provided in App. \ref{app:pos_encoding}.

\subsection{Training Objectives}
\textbf{Unified training.}
To jointly optimize tracking and grounding within a single model, training follows a unified hybrid sampling scheme, drawing mini-batches from the three data sources (tracking, joint tracking and grounding) with probabilities $[0.45, 0.40, 0.15]$.
Task decoupling is achieved by dynamic input masking under the same architecture.
For grounding samples, the template tokens in the input sequence $S_{\mathrm{in}}$ are explicitly masked, preventing appearance-based cues and forcing language-only localization from scratch.
A substantial fraction of template-only tracking data is included during training, which empirically mitigates language-dominated drift and strengthens the model’s ability to track based on the visual template.
\begin{figure}[hbt]
\centering
\includegraphics[width=2.8in]{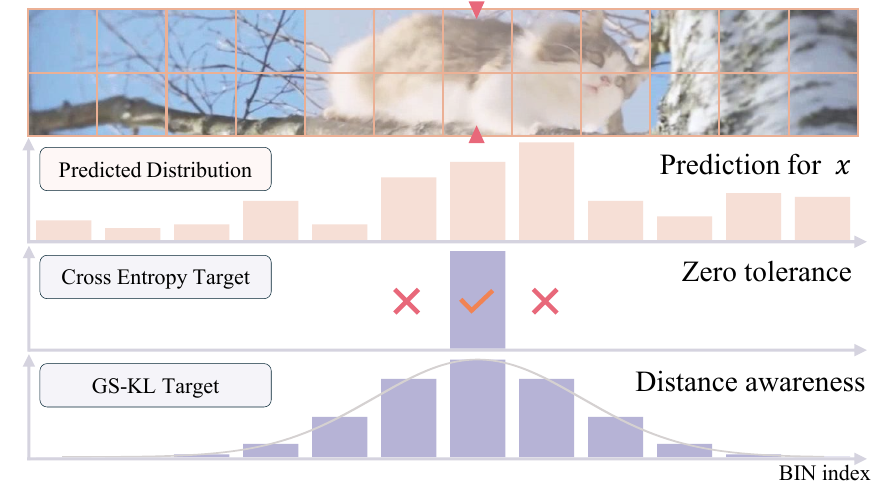}
\hfil
\vspace{-1ex}
\caption{Comparison between cross-entropy and the proposed GS-KL loss for discretized coordinate prediction. GS-KL replaces the one-hot target with a Gaussian-smoothed soft target, assigning higher probability mass to bins closer to the ground truth and making token classification distance-aware.}
\vspace{-2ex}
\label{loss}
\end{figure}

\textbf{Gaussian-smoothed KL loss.}
Autoregressive localization discretizes continuous coordinates into categorical tokens, but standard cross-entropy ignores the ordinal structure of coordinate bins and penalizes near-miss and far-off predictions equally \cite{ref51}.
We therefore propose a Gaussian-smoothed KL (GS-KL) loss, which replaces the one-hot target with a distance-aware soft distribution over neighboring bins (Fig. \ref{loss}).
For each coordinate head $i\in\{1,2,3,4\}$ with ground-truth bin $y_i$, the target distribution is defined as:

\begin{equation}
T_i(k)=\frac{\exp\!\left(-\frac{(k-y_i)^2}{2\sigma^2}\right)}{\sum_{j}\exp\!\left(-\frac{(j-y_i)^2}{2\sigma^2}\right)} .
\end{equation}
where $\sigma$ controls the smoothing radius.
Given the predicted distribution $P_i$, the localization loss is:

\begin{equation}
\mathcal{L}_{\mathrm{loc}}=\sum_{i=1}^{4} D_{\mathrm{KL}}\!\left(T_i \,\|\, P_i\right).
\end{equation}
By assigning larger supervision weights to bins closer to the ground truth, GS-KL injects spatial ordinality into token classification, alleviates quantization ambiguity, and provides regression-like localization sensitivity.
The overall objective is:

\begin{equation}
\mathcal{L}=\mathcal{L}_{\mathrm{loc}}+\lambda_{\mathrm{ratio}} \mathcal{L}_{\mathrm{ratio}}.
\end{equation}

\begin{table*}[t]
\renewcommand{\arraystretch}{1.1}
\setlength\tabcolsep{5.2pt}
\small
\centering
\caption{Comparison of our method with state-of-the-art approaches on TNL2k, LaSOT and OTB99 datasets. The best and second-best results are highlighted in bold and underlined respectively.}
\begin{tabular}{c|c|ccc|ccc|ccc}
\toprule
\multirow{2}{*}{Tracker} & \multirow{2}{*}{Initialize} & \multicolumn{3}{c|}{TNL2K} & \multicolumn{3}{c|}{LaSOT} & \multicolumn{3}{c}{OTB99} \\ \cline{3-11}
& & AUC & Prec & NPrec & AUC & Prec & NPrec & AUC & Prec & NPrec \\ \hline
TNL2K-II~\cite{ref23} & NL+BBOX & 41.7 & 42.0 & 50.0 & 51.0 & 55.0 & - & 68.0 & 88.0 & - \\
JointNLT~\cite{ref1} & NL+BBOX & 56.9 & 58.1 & 73.6 & 60.4 & 63.6 & 69.4 & 65.3 & 85.6 & 79.5 \\
QueryNLT~\cite{ref12} & NL+BBOX & 57.8 & 58.7 & 75.6 & 59.9 & 63.5 & 69.6 & 66.7 & 88.2 & 82.4 \\
UVLTrack-B~\cite{ref2} & NL+BBOX & 63.1 & 66.7 & - & 69.4 & 75.9 & - & 69.3 & 89.9 & - \\
DUTrack-256~\cite{ref3} & NL+BBOX & 64.9 & 70.6 & 82.9 & \underline{73.0} & \underline{81.1} & 83.8 & 70.9 & 93.9 & - \\
DUTrack-384~\cite{ref3} & NL+BBOX & 65.6 & \textbf{71.9} & 83.2 & \textbf{74.1} & \textbf{82.9} & \underline{84.9} & 71.3 & \textbf{95.7} & - \\
MambaVLT~\cite{ref27} & NL+BBOX & \textbf{66.5} & 69.9 & \textbf{90.9} & 66.6 & 71.0 & 77.3 & \underline{72.2} & 94.4 & 88.1 \\
SAVLT-B~\cite{ref34} & NL+BBOX & \underline{66.4} & \underline{71.4} & 84.3 & \underline{73.0} & 80.5 & \textbf{85.9} & 70.4 & 92.2 & - \\
\rowcolor{ourscolor} \textbf{LVTrack-256} & NL+BBOX & 64.7 & 68.8 & 84.9 & 69.6 & 76.0 & 81.3 & 71.7 & 94.8 & \underline{88.7} \\ 
\rowcolor{ourscolor} \textbf{LVTrack-384} & NL+BBOX & \textbf{66.5} & 70.7 & \underline{86.8} & 72.7 & 78.5 & 82.9 & \textbf{72.4} & \underline{95.3} & \textbf{88.9} \\ \hline
TNL2K-I~\cite{ref23} & NL & 11.4 & 6.4 & 11.0 & 51.1 & 49.3 & - & 19.0 & 24.0 & - \\
CTRNLT~\cite{ref7} & NL & 14.0 & 9.0 & - & 52.0 & 51.0 & - & 53.0 & 72.0 & - \\
JointNLT~\cite{ref1} & NL & 54.6 & 55.0 & 70.6 & 56.9 & 59.3 & 64.5 & 59.2 & 77.6 & - \\
QueryNLT~\cite{ref12} & NL & 53.3 & 53.0 & 70.4 & 54.2 & 55.0 & 62.5 & 61.2 & 81.0 & 73.9 \\
UVLTrack-B~\cite{ref2} & NL & 55.7 & 57.2 & - & 57.2 & 61.0 & - & 60.1 & 79.1 & - \\
MambaVLT~\cite{ref27} & NL & \underline{58.4} & 58.9 & \textbf{80.9} & 55.8 & 57.2 & 63.7 & 58.9 & 79.2 & 72.0 \\
SAVLT-B~\cite{ref34} & NL & 57.1 & \underline{59.8} & 73.1 & \textbf{62.8} & \textbf{67.9} & \textbf{73.6} & 61.7 & 81.2 & - \\
\rowcolor{ourscolor} \textbf{LVTrack-256} & NL & 57.6 & 58.1 & 76.7 & 58.7 & 61.9 & 68.3 & \underline{63.3} & \underline{82.9} & \underline{75.9} \\
\rowcolor{ourscolor} \textbf{LVTrack-384} & NL & \textbf{58.7} & \textbf{59.9} & \underline{79.4} & \underline{61.3} & \underline{64.5} & \underline{71.2} & \textbf{63.6} & \textbf{83.0} & \textbf{76.3} \\
\bottomrule
\end{tabular}
\vspace{-1ex}
\label{tab:table2}
\end{table*}

%% file: sec/5_experiments.tex
\section{Experiments}

\subsection{Implementation Details}

\textbf{Network configuration.}
We adopt PE-Base \cite{ref28} as the backbone and keep all its parameters frozen.
For the MGFI, modules are inserted into the visual layers $\{5,6,7,8,9,10\}$, each injecting language features from layers $\{14,16,18,20,22,24\}$, respectively.
The default input resolutions for the search and template images are $256\times256$ and $128\times128$, while the high-resolution variant uses $384\times384$ and $192\times192$, respectively.
Both the tracking encoder and decoder use an embedding dimension of $D=256$, with $8$ and $4$ layers, respectively.
We use $64$ memory tokens, and set the decoder discretization to $\texttt{BINS}=4000$.

\textbf{Training details.}
We train on the training splits of OTB99 \cite{ref21}, LaSOT \cite{ref31}, TNL2K \cite{ref23}, RefCOCOg \cite{ref32}, and GOT-10k \cite{ref33}, and evaluate on the official validation/test splits of the first four datasets.
Training is performed with AdamW \cite{ref47} using a learning rate of $2\times10^{-4}$ and a batch size of $192$, following a cosine-annealing schedule with $20$ warmup epochs for a total of $80$ epochs.
Hyperparameters are set to $\sigma=4$ for GS-KL, $\rho=0.6$ for MGFI decay factor, $\lambda_{\text{ratio}}=0.01$ for the amplitude ratio loss, and $\tau_{\mathrm{mem}}=0.03$ for memory update.
More discussion on hyperparameters is provided in App. \ref{app:hyperparameter}.

\textbf{Evaluation metrics.}
Following standard protocols, we report AUC, Precision (Prec), and Normalized Precision (NPrec) for tracking, measuring overlap quality and center-based localization accuracy.
For grounding, we report Top-1 accuracy on RefCOCOg val set at an IoU threshold of 0.5.

\subsection{Efficiency Analysis}

\begin{table}[ht]
\small
\setlength\tabcolsep{1.5pt}
\centering
\caption{Comparison of training budget and efficiency.}
\begin{tabular}{lcccc}
\toprule
\textbf{Metric} & \textbf{JointNLT} & \textbf{UVLTrack} & \textbf{SAVLT} & \textbf{Ours} \\
\midrule
Trainable parameters & 153M & 169M & 129M & \textbf{100M} \\
Training epochs           & 300  & 300  & 300  & \textbf{80}   \\
Samples per epoch& 60K& 30K& 60K& \textbf{30K}\\
\bottomrule
\end{tabular}
\vspace{-2ex}
\label{tab:table1}
\end{table}
Training efficiency is a core advantage of our architecture.
By leveraging a frozen VLP model, we avoid the additional vision–language alignment fine-tuning stage commonly required in prior work, and keep a large fraction of parameters fixed.
Tab.~\ref{tab:table1} summarizes the resulting savings.
LVTrack optimizes only 100M parameters and reaches its best performance within 80 epochs, whereas strong baselines such as SAVLT and UVLTrack are typically trained for up to 300 epochs.
Combined with a lighter schedule of 30K samples per epoch, this substantially reduces the total number of optimization steps and overall compute.
In our setup (8 NVIDIA RTX 4090 GPUs, bfloat16), the LVTrack-256 model converges in approximately 2 hours.
Meanwhile, LVTrack-256 runs at about 40 FPS for tracking (22 FPS for LVTrack-384), comparable to prior methods.

\subsection{Comparison with State-of-the-art}

LVTrack is compared with recent vision-language trackers under two initialization protocols (Tab.~\ref{tab:table2}): NL (language-only initialization) and NL+BBOX (language plus a first-frame box).

\textbf{NL-initialized performance.}
NL-only tracking requires localizing the object in the first frame purely from the referring expression, thus directly reflecting grounding quality and its coupling with temporal tracking.
LVTrack-384 achieves the best AUC and Prec on TNL2K (58.7/59.9), the best results on OTB99 across all metrics (63.6/83.0/76.3), and the second-best results on LaSOT (61.3/64.5/71.2).
Compared with strong NL trackers such as MambaVLT and SAVLT-B, LVTrack shows stronger overall performance across datasets, and its 256-resolution variant also remains highly competitive, ranking second on OTB99 (63.3/82.9/75.9) and achieving 57.6 AUC on TNL2K.
Although SAVLT-B obtains the highest LaSOT scores, LVTrack delivers more balanced performance across datasets.
These results show that LVTrack effectively transfers pretrained knowledge to language-only initialization and maintains robust temporal localization in complex scenarios.

\textbf{NL+BBOX-initialized performance.}
When a visual template is available, appearance cues dominate and language mainly serves as auxiliary semantics.
LVTrack-384 remains competitive with specialized NL+BBOX trackers, obtaining the best AUC on TNL2K (66.5), the best AUC and NPrec on OTB99 (72.4/88.9), and the second-best OTB99 Prec (95.3), while LVTrack-256 also maintains strong performance.
On LaSOT, DUTrack performs best under this protocol, benefiting from an NL+BBOX-specific design, but it does not support NL-only grounding-style initialization.
In contrast, LVTrack achieves competitive NL+BBOX performance while retaining strong NL-initialized tracking ability, showing that the proposed dual-mode gating can exploit the visual template without discarding useful language cues.
Together with its much lower training cost, these results demonstrate the effectiveness and practicality of LVTrack across different initialization protocols.

\begin{table}[h]
\small
\setlength\tabcolsep{1.7pt}
\caption{Visual grounding comparison on the RefCOCOg validation set.}
\begin{tabular}{lccccc}
\toprule
Method & VLTVG & JointNLT & QueryNLT & UVLTrack & \textbf{Ours} \\
\midrule
Acc. & 72.98 & 70.07 & 72.00 & 73.86 & \textbf{74.88} \\
\bottomrule
\end{tabular}
\label{tab:table3}
\end{table}
\textbf{Visual grounding performance.}
To further assess pure language understanding and object localization, we compare on the RefCOCOg validation set.
Visual grounding requires localizing the referred object in a single image solely from text, serving as the basis of NL-initialized tracking and a key indicator of multimodal alignment quality.
As shown in Tab. \ref{tab:table3}, LVTrack-256 achieves 74.88\% accuracy, surpassing all competitors.
Notably, this result is obtained using the same unified model deployed for tracking, rather than re-training a dedicated grounding model.
This confirms stronger language understanding in our architecture and shows that MGFI constrains language reliance during tracking without degrading grounding capability.

\begin{table}[t]
\centering
\small
\renewcommand{\arraystretch}{1.1}
\setlength{\tabcolsep}{3.2pt}
\caption{Ablation study on LaSOT (AUC), OTB99 (AUC), and RefCOCOg (Accuracy). Tracking is evaluated under the NL-initialized setting.}
\begin{tabular}{lccc}
\toprule
Model & LaSOT & OTB99 & RefCOCOg \\
\hline
LVTrack & \textbf{58.7} & \textbf{63.3} & \textbf{74.88} \\
\hline
w/o GS-KL Loss & 56.4 & 59.6 & 70.92 \\
w/o Amplitude Ratio Loss & 58.3 & 60.2 & 71.46 \\
w/o Dual Mode Fusion & 57.7 & 58.7 & 71.14 \\
w/o Memory Fusion & 57.3 & 60.3 & 73.70 \\
w/o 2D RoPE & 55.3 & 58.1 & 71.72\\
\bottomrule
\end{tabular}
\label{tab:ablation}
\end{table}
\subsection{Ablation Study}

We conduct ablations on LaSOT, OTB99, and RefCOCOg. All tracking results are reported under the NL-initialized setting with LVTrack-256. Detailed results are summarized in Tab.~\ref{tab:ablation}.

\begin{figure}[ht]
\centering
\includegraphics[width=2.8in]{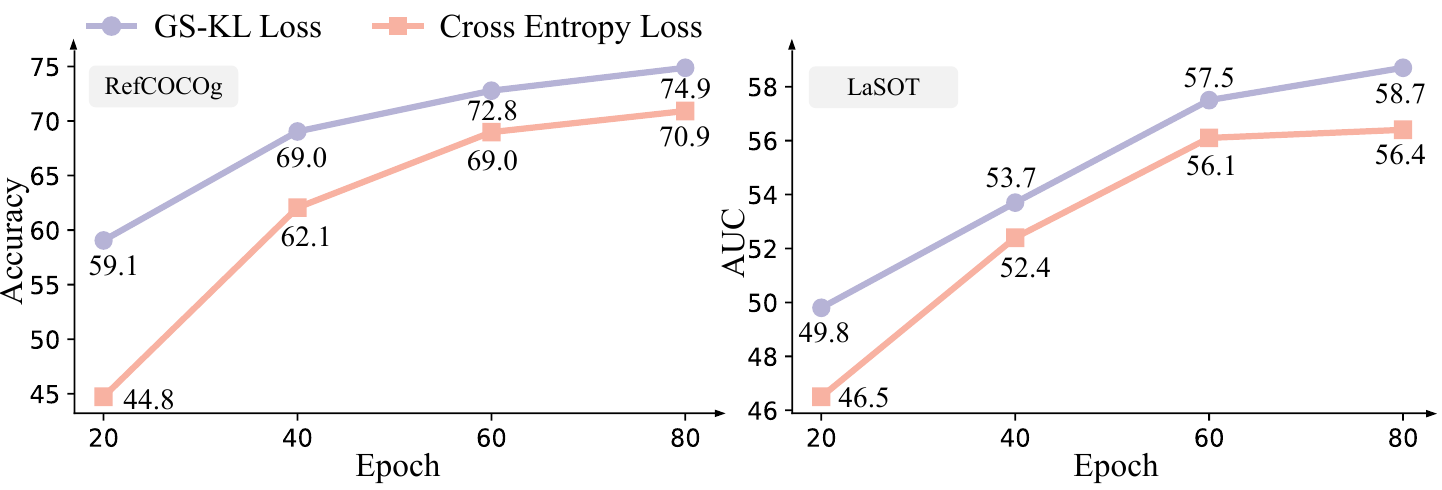}
\hfil
\caption{Comparison of GS-KL loss and cross-entropy loss across training epochs. The left panel reports grounding accuracy on RefCOCOg, and the right panel reports NL-initialized tracking performance on LaSOT.}
\label{curve}
\end{figure}

\textbf{GS-KL loss.}
We compare the proposed GS-KL loss with standard cross-entropy (CE). As shown in Tab.~\ref{tab:ablation}, removing GS-KL consistently degrades performance across all benchmarks. The training dynamics in Fig.~\ref{curve} further indicate both a higher asymptotic accuracy and faster convergence: GS-KL yields more stable optimization and opens a larger margin over CE in early training. This validates that, under next-token prediction for localization, distance-aware soft targets effectively mitigate quantization error induced by coordinate discretization, enabling pixel-level fine-grained localization rather than coarse categorical classification.
\begin{figure*}[t]
\centering
\includegraphics[width=6in]{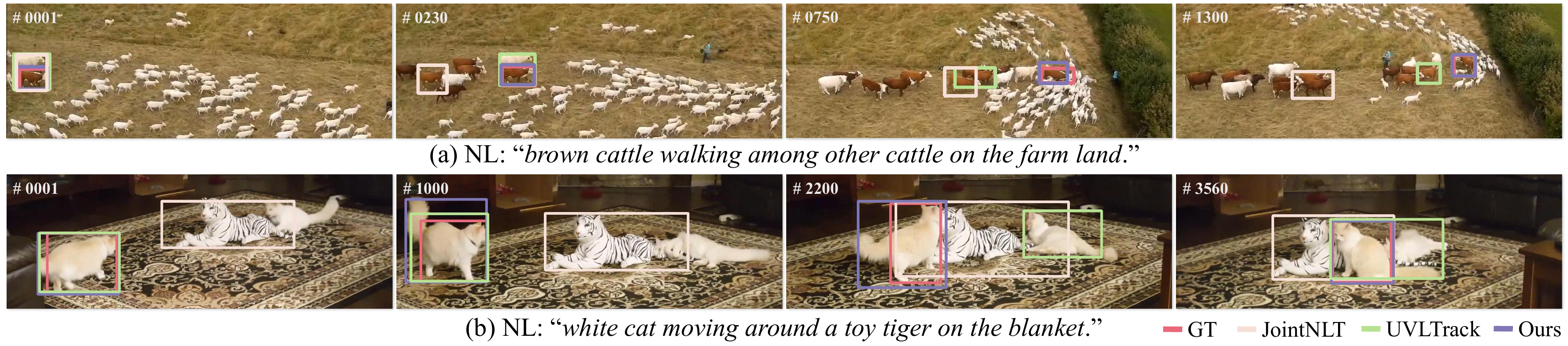}
\hfil
\vspace{-1ex}
\caption{Tracking visualizations under NL initialization. (a) shows the \textit{cattle-13} sequence and (b) shows the \textit{cat-3} sequence from LaSOT.}
\label{vis}
\vspace{-2ex}
\end{figure*}
\textbf{MGFI.}
The injection module is designed to reconcile language-sensitive grounding with vision-anchored tracking.
Removing the amplitude ratio loss consistently degrades LaSOT/OTB99/RefCOCOg by -0.4/-3.1/-3.42, with the larger drop on OTB99 indicating increased susceptibility to similar distractors.
This verifies that regulating tracking-mode injection suppresses language-dominated drift while preserving useful textual guidance.
Further removing dual-mode fusion causes larger losses (-1.0/-4.6/-3.74), confirming that mode-specific fusion is necessary to calibrate textual injection under different language reliance.

\begin{figure}[h]
\centering
\includegraphics[width=2.8in]{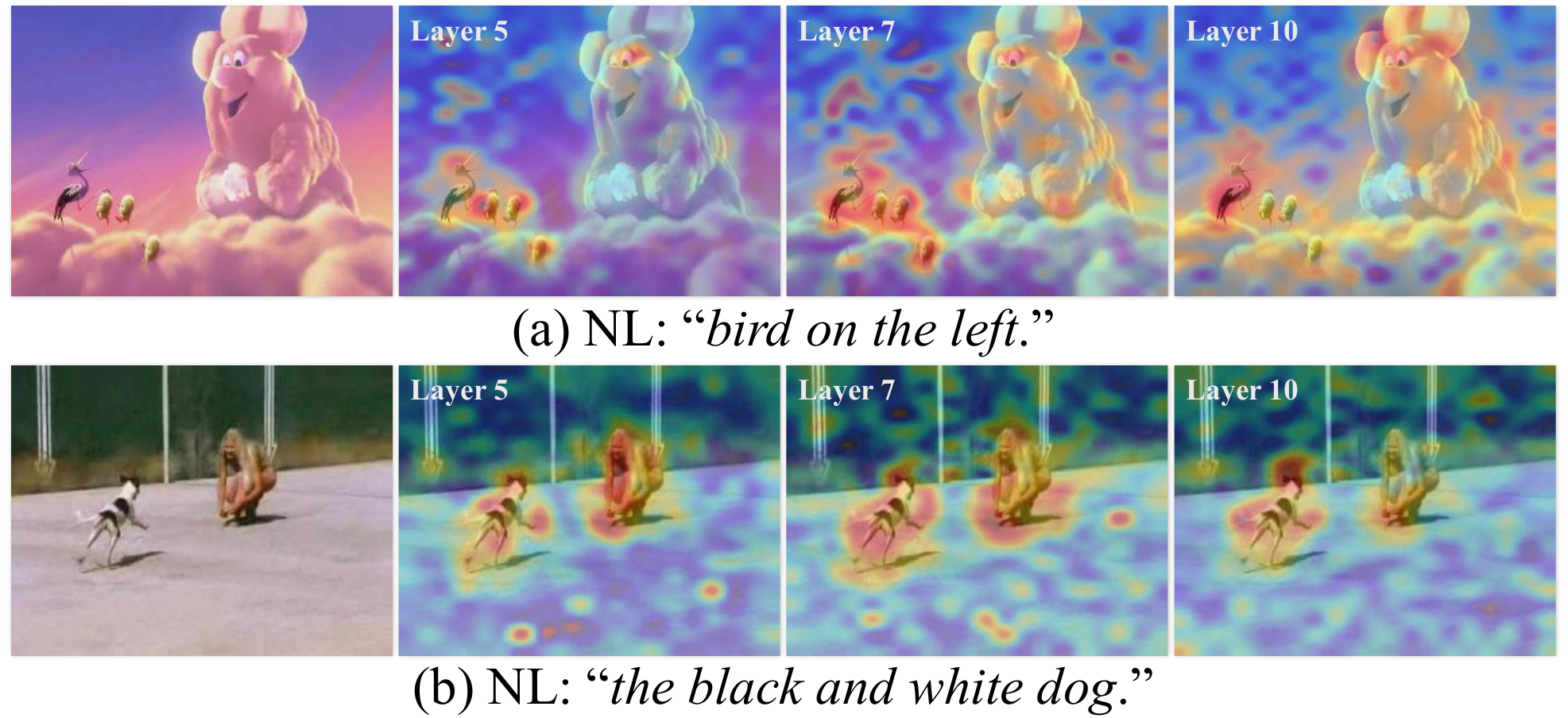}
\hfil
\vspace{-1ex}
\caption{Feature map visualizations as language is progressively injected. The panel includes the input image and feature-magnitude heatmaps from layers 5, 7, and 10 of the visual backbone.}
\label{feature}
\vspace{-1ex}
\end{figure}

Fig.~\ref{feature} visualizes this modulation effect.
As MGFI is progressively applied across layers, activations evolve from diffuse salient regions to expression-relevant target regions, showing that MGFI sharpens spatial selectivity and reduces distractor responses.

\textbf{Memory fusion.}
The temporal memory module improves feature robustness by leveraging historical information. It mainly affects long-term tracking, with AUC drops of 1.4 and 3 points on LaSOT and OTB99, respectively, while having a smaller impact on grounding. This suggests that a dynamically updated appearance memory effectively handles deformation and occlusion, strengthening persistent localization in complex temporal scenarios.

\textbf{Positional embedding.}
Removing 2D RoPE from both the encoder and decoder (using only learnable absolute embeddings) yields a consistent and notable drop on all benchmarks (Tab.~\ref{tab:ablation}), suggesting that absolute positions alone are insufficient to capture relative geometry and cross-frame transformations.
In contrast, 2D RoPE injects relative inductive bias into attention, improving spatial topology preservation and localization stability.
It is also more extensible to resolution changes, particularly when template and search have mismatched sizes.
Notably, our backbone (PE) already adopts RoPE, and keeping 2D RoPE in the downstream encoder-decoder maintains positional-code consistency and better inherits its geometric prior.

\textbf{Qualitative results.}
Fig. \ref{vis} compares three representative referring trackers. 
In sequence (a), LVTrack is the only method that precisely localizes the described ``brown cattle'' in the first-frame grounding stage; JointNLT and UVLTrack produce boxes that partially cover the object but also include distractors. 
As clutter increases in subsequent frames, both baselines drift, whereas LVTrack remains locked onto the object. 
In sequence (b), only UVLTrack and LVTrack initialize correctly, yet UVLTrack later switches to a similar cat and, under overlap, fails to isolate a single instance, enclosing both cats in one box. 
LVTrack consistently tracks the correct individual throughout.

%% file: sec/6_conclusion.tex
\section{Conclusion}
We present LVTrack, an efficient referring single-object tracking framework that unifies grounding and tracking in an autoregressive formulation.
A mode-conditioned gated feature injector adaptively calibrates textual guidance, enabling language-sensitive grounding while mitigating language-dominated drift in tracking.
Together with GS-KL supervision, hybrid positional encoding, and appearance memory, LVTrack improves localization accuracy and temporal robustness.
These designs make it possible to directly leverage a frozen vision-language pretrained backbone for RSOT, achieving strong benchmark performance with substantially reduced training cost.

%% file: sec/7_appendix.tex
\appendix

\section{Inference Details}
\label{app:inference}

At inference, LVTrack adopts lightweight robustness heuristics to mitigate semantic drift and stabilize autoregressive tracking.
Under NL initialization, the first frame is localized in grounding mode from the referring expression to obtain the initial box and build the template; this step is skipped when the initial bounding box is provided.
Subsequent frames are processed in tracking mode within a local search region using autoregressive decoding.

During inference, memory updates are controlled by prediction confidence to avoid refreshing the appearance memory with unreliable target features.
Specifically, for each coordinate head in $\{x,y,w,h\}$, we take the top-1 softmax probability as the confidence score, denoted as $\{c_x,c_y,c_w,c_h\}$.
The memory is updated only when the aggregated confidence is sufficiently high, which provides a simple reliability criterion for template refresh and reduces error accumulation under occlusion or distractors.

In addition, spatial directive words in the expression (e.g., up/left/center) typically describe only the first-frame location and may mislead temporal tracking.
Thus, we keep the original expression $L^{g}$ only for first-frame grounding, and apply directive stripping to obtain $L^{t}=\mathcal{R}(L^{g})$ for subsequent frames, re-encoding language features for MGFI.
The same preprocessing is used during training.
Finally, following SeqTrack, we apply a Hanning-window penalty to the center-coordinate distribution to bias predictions toward the search-window center, suppressing spurious large jumps \cite{ref6, ref48}.

\section{Hyperparameter Analysis}
\label{app:hyperparameter}

\begin{table*}[t]
\small
\centering
\renewcommand{\arraystretch}{1.1}
\setlength{\tabcolsep}{16pt}
\caption{Hyperparameter analysis on LaSOT (AUC), OTB99 (AUC), and RefCOCOg (Accuracy). Tracking is evaluated under the NL-initialized setting. Data sampling ratio is reported as a three-number tuple, denoting the proportions of tracking, joint tracking, and grounding samples, respectively.
}
\begin{tabular}{l c c c c}
\toprule
Experiment & Parameter & LaSOT & OTB99 & RefCOCOg \\
\midrule
\multirow{3}{*}{ViT layers}
 & 9  & 56.3 & 60.8 & 71.18 \\
 & 10 & \textbf{58.7} & \textbf{63.3} & \textbf{74.88} \\
 & 11 & 56.6 & 58.7 & 71.64 \\
\midrule
\multirow{3}{*}{Decay factor $\rho$ in $\mathcal{L}_{\text{ratio}}$}
 & 0.5 & 57.6 & 60.4 & 74.20 \\
 & 0.6 & \textbf{58.7} & \textbf{63.3} & \textbf{74.88} \\
 & 0.7 & 58.2 & 61.1 & 73.88 \\
\midrule
\multirow{3}{*}{Bandwidth $\sigma$ in $\mathcal{L}_{\mathrm{loc}}$}
 & 2 & 57.3 & 62.3 & 74.66 \\
 & 4 & \textbf{58.7} & \textbf{63.3} & \textbf{74.88} \\
 & 6 & 57.1 & 61.2 & 74.48 \\
\midrule
\multirow{3}{*}{Data sampling ratio}
 & [0.40, 0.50, 0.10] & 58.3 & 61.6 & 72.12 \\
 & [0.45, 0.40, 0.15] & \textbf{58.7} & \textbf{63.3} & 74.88 \\
 & [0.40, 0.35, 0.25] & 57.6 & 61.4 & \textbf{74.94} \\
\bottomrule
\end{tabular}
\label{tab:hyper}
\end{table*}

\textbf{ViT layer selection.}
LVTrack extracts visual embeddings from an intermediate layer of the frozen VLP ViT rather than the final layer.
This is because CLIP-style contrastive pretraining primarily optimizes global image-text alignment: in the last few transformer blocks, information is progressively compressed into a global representation, and emerging ``global'' tokens attract attention from most patch tokens, making the network behave like a global-information decoder \cite{ref28, ref50}.
While beneficial for classification-level semantics, this late-stage aggregation weakens patch-level distinctiveness and disrupts spatial correspondence, which is detrimental to tracking and other localization-sensitive tasks that rely on stable local geometry.
In this ablation, we isolate the effect of the ViT output depth by changing \emph{only} the tapped backbone layer and keeping all other components and hyperparameters fixed; the MGFI stack uses the same number of modules and is shifted accordingly to stay aligned with the selected output depth.
Tab. \ref{tab:hyper} confirms this behavior: using the 10th layer (out of 12) yields the best overall performance across both tracking and grounding (e.g., 58.7/63.3 AUC on LaSOT/OTB99 and 74.88 accuracy on RefCOCOg), outperforming a shallower (9th) or deeper (11th) choice.
Overly shallow output features lack semantic understanding, even though they may preserve stronger spatial cues. We therefore choose layer 10 as the default output, as it provides a better balance between grounding and tracking performance.

\textbf{Decay factor $\rho$ in the amplitude ratio loss.}
The amplitude ratio loss explicitly enforces a \emph{vision-first, language-secondary} fusion hierarchy by constraining the tracking-mode gate magnitude to be approximately a scaled version of the grounding-mode magnitude, i.e., $\lambda^{\mathcal{T}} \approx \rho \cdot \lambda^{\mathcal{G}}$. 
Tab. \ref{tab:hyper} shows that $\rho=0.6$ achieves the best balance across tasks.
When $\rho$ is smaller (0.5), the constraint is overly strict, suppressing linguistic guidance too aggressively. This slightly harms grounding and also weakens NL-initialized tracking, where first-frame localization quality is critical.
When $\rho$ is larger (0.7), the constraint becomes too loose, allowing stronger language influence during tracking and increasing susceptibility to language-dominated drift, which is reflected by a noticeable drop on OTB99 (63.3 $\rightarrow$ 61.1).
Notably, relaxing the amplitude constraint also reduces grounding performance.
We attribute this to the training dynamics on tracking samples, where ambiguous language cues that are insufficiently constrained act as distractors and encourage the model to down-weight linguistic signals.
Since tracking and grounding share the same cross-attention layer, this reduced reliance on language transfers to grounding and degrades its accuracy.

\textbf{Bandwidth $\sigma$ in the GS-KL localization loss.}
We further study the bandwidth $\sigma$ in the GS-KL objective, which controls the tolerance radius on the discrete bin lattice and thus the strength of distance-aware supervision.
As reported in Tab. \ref{tab:hyper}, $\sigma=4$ achieves the best overall performance on both tracking and grounding, outperforming smaller ($\sigma=2$) and larger ($\sigma=6$) bandwidths.
With a small bandwidth ($\sigma=2$), the target distribution becomes sharply peaked and the objective approaches a cross-entropy-like regime, offering limited tolerance for near-miss bins and thus weaker distance-aware supervision, which exacerbates discretization ambiguity.
With a large bandwidth ($\sigma=6$), the target is overly diffuse, weakening fine-grained corrective gradients and hurting precise coordinate refinement, reflected by drops in both tracking AUC and grounding accuracy.
Overall, $\sigma=4$ provides the best trade-off between locality and sharpness, and is adopted as the default.

\textbf{Unified sampling ratios.}
We adopt a unified hybrid sampling scheme over three data sources (tracking, joint tracking, grounding).
The data sampling ratio is crucial for training our model and serves as an effective mechanism to balance grounding and tracking.
A proper proportion of template-only tracking samples mitigates language-dominated drift during tracking, whereas an excessive grounding ratio can exacerbate this issue.
Tab. \ref{tab:hyper} indicates that the default ratio $[0.45, 0.40, 0.15]$ provides the strongest tracking performance while maintaining high grounding accuracy.
Reducing the grounding proportion to $0.10$ ($[0.40, 0.50, 0.10]$) degrades RefCOCOg accuracy (72.12) and also hurts NL-initialized tracking, consistent with the fact that NL-only tracking hinges on first-frame grounding quality and its coupling with temporal tracking.
Conversely, increasing the grounding proportion to $0.25$ ($[0.40, 0.35, 0.25]$) yields a marginal gain on RefCOCOg (74.94) but reduces tracking AUC (e.g., LaSOT 58.7 $\rightarrow$ 57.6), suggesting that over-emphasizing grounding can bias optimization toward language-dominant cues and weaken the tracker’s reliance on visual template consistency.
Overall, $[0.45, 0.40, 0.15]$ is a robust operating point that best preserves the intended tracking-grounding trade-off under unified training.

\section{Backbone Selection Rationale}
\label{app:backbone}

\begin{table*}[ht]
\centering
\small
\renewcommand{\arraystretch}{1.1}
\setlength{\tabcolsep}{8pt}
\caption{Backbone comparison under an identical LVTrack setting on LaSOT (AUC), OTB99 (AUC), and RefCOCOg (Accuracy). We replace only the frozen VLP backbone (all base-scale) while keeping the architecture and training protocol unchanged.}
\begin{tabular}{l c c c c c c c}
\toprule
\multirow{2}{*}{Backbone} &
\multicolumn{1}{c}{RefCOCOg} &
\multicolumn{3}{c}{LaSOT} &
\multicolumn{3}{c}{OTB99} \\
\cmidrule(lr){2-2}\cmidrule(lr){3-5}\cmidrule(lr){6-8}
& Accuracy & AUC & Prec & NPrec & AUC & Prec & NPrec \\
\midrule
PE \cite{ref28} & \textbf{74.88} & \textbf{58.7} & \textbf{61.9} & \textbf{68.3} & \textbf{63.3} & \textbf{82.9} & \textbf{75.9} \\
SigLIP 2 \cite{ref43} & 70.35 & 56.9 & 59.7 & 66.5 & 62.1 & 81.6 & 74.1 \\
CLIP \cite{ref10} & 70.86 & 56.7 & 58.8 & 66.0 & 61.6 & 80.3 & 73.7 \\
\bottomrule
\end{tabular}
\label{tab:backbone}
\end{table*}

A key factor behind LVTrack's strong performance is its use of a vision-language pretrained model.
In this section, we compare different VLP backbones within LVTrack and justify our final choice of PE as the backbone.  
Tab.~\ref{tab:backbone} reports results for three base-scale VLP models under an identical LVTrack pipeline, where we replace only the backbone while keeping the architecture, training schedule, and all hyperparameters fixed.  
Because PE uses a 24-layer text encoder while SigLIP 2 and CLIP use 12 layers, we align MGFI by injecting language features from the last six text layers, namely $[14,16,18,20,22,24]$ for PE and $[7,8,9,10,11,12]$ for SigLIP 2/CLIP.  
All three backbones share a 12-layer visual encoder, and the visual injection depths are kept consistent across models.  
Under this controlled setting, PE yields the best performance across both grounding and tracking.  

We believe PE performs better mainly for the following reasons.
First and most importantly, PE is originally trained with progressive resolutions and 2D RoPE, which is crucial because our template and search regions have different sizes while the backbone remains frozen.
The relative positional modeling of RoPE helps maintain spatial consistency across resolutions, improving robustness to resolution changes.
Second, its deeper 24-layer text encoder offers stronger semantic parsing and compositional understanding, which is evidenced by the clear gain on RefCOCOg and yields more reliable NL-only initialization for tracking.
Third, PE adopts improved training strategies and a stronger data engine, leading to higher robustness and better generalization under long-term tracking challenges such as occlusion, distractors, and appearance variation.
This is particularly important for LVTrack, where the same frozen backbone must process template and search at different resolutions.
Moreover, our encoder-decoder also uses 2D RoPE, and choosing a backbone that already employs RoPE makes the overall positional modeling more consistent, which empirically leads to better performance.
In contrast, CLIP and SigLIP2 mainly rely on absolute position embedding interpolation without RoPE, and typically exhibit weaker multi-resolution adaptability.

However, the advantage of LVTrack should not be viewed as a direct benefit of using PE as the backbone.
In our preliminary experiments, we also applied PE to prior feature-concatenation-based trackers. Specifically, we replaced both their visual backbone and BERT-based text encoder with PE, while keeping the original pipeline of concatenating visual and textual features before feature modeling. After retraining, these models showed degraded tracking performance.
This suggests that frozen VLP representations are not naturally aligned with the requirements of RSOT.
LVTrack addresses this issue by introducing MGFI, where mode-conditioned dual injection applies controllable feature-map perturbations to bridge this gap.
Along with the use of shallower ViT features and task-oriented optimization designs, LVTrack enables frozen VLP models to be effectively incorporated into RSOT.

\section{MGFI Perturbation Analysis}
\label{app:mgfi_analysis}

This section provides the perturbation bound for MGFI.
Language injection modifies the visual feature consumed by the downstream predictor.
Let $z=h(f)$ denote the decoder outputs as a function of the visual feature $f$.
With $f \triangleq f_n^s$ and $\Delta^{\mathcal{T}} \triangleq \Delta_n^{\mathcal{T}}$, and assuming $h$ is locally $L_h$-Lipschitz, the joint-tracking-mode deviation is bounded by the perturbation magnitude:

\begin{equation}
\begin{aligned}
\big\|z^{\mathcal{T}}-z^{\text{no-inj}}\big\|
&= \big\|h(f+\Delta^{\mathcal{T}})-h(f)\big\| \\
&\le L_h\,\big\|\Delta^{\mathcal{T}}\big\|.
\end{aligned}
\label{eq:lipschitz_drift}
\end{equation}
For brevity, $A_n$ denotes the MHCA residual at layer $n$ computed from $(f_n^s, f^L_{\tau(n)})$, and $M_n^{m}$ denotes the mode-$m$ MLP residual at the same layer.
By triangle inequality,
\begin{equation}
\begin{aligned}
\big\|\Delta_n^{\mathcal{T}}\big\|
&\le
\big\|\lambda_{\text{attn}}^{\mathcal{T}} \odot A_n\big\|
+
\big\|\lambda_{\text{mlp}}^{\mathcal{T}} \odot M_n^{\mathcal{T}}\big\| \\
&\le
\big\|\lambda_{\text{attn}}^{\mathcal{T}}\big\|_{\infty}\,\big\|A_n\big\|
+
\big\|\lambda_{\text{mlp}}^{\mathcal{T}}\big\|_{\infty}\,\big\|M_n^{\mathcal{T}}\big\|,
\end{aligned}
\label{eq:gate_bound}
\end{equation}
which shows that the gate norms directly upper-bound the maximum impact of language on tracking decisions.

\section{Positional Encoding and Position Modeling Details}
\label{app:pos_encoding}

In LVTrack, to effectively capture the geometric structure of visual objects and establish precise spatial correspondence between the template and the search region, we propose a hybrid positional encoding scheme. This scheme combines a shared 2D absolute positional encoding ($\mathcal{P}_{\mathrm{abs}}$) with a 2D axial rotary positional embedding ($\mathcal{P}_{\mathrm{RoPE}}$). This section details the mathematical formulation and implementation, followed by a theoretical analysis of its advantages in visual tracking tasks.

\begin{table}[ht]
\small
\centering
\renewcommand{\arraystretch}{1.1}
\setlength{\tabcolsep}{2.4pt}
\caption{Ablation results of using different forms of positional encoding on RefCOCOg, LaSOT, and OTB99. $\mathrm{1D}\ \mathcal{P}_{\mathrm{abs}}$ denotes a single 1D absolute positional embedding table with length equal to the total token sequence, i.e., the positional coordinates of template and search tokens are \emph{not} defined under a shared coordinate system. $\mathrm{1D}\ \mathcal{P}_{\mathrm{RoPE}}$ is defined analogously.}
\begin{tabular}{l c c c c c}
\toprule
\multirow{2}{*}{Positional Encoding} &
\multicolumn{1}{c}{RefCOCOg} &
\multicolumn{2}{c}{LaSOT} &
\multicolumn{2}{c}{OTB99} \\
\cmidrule(lr){2-2}\cmidrule(lr){3-4}\cmidrule(lr){5-6}
& Accuracy & AUC & Prec & AUC & Prec \\
\midrule
2D $\mathcal{P}_{\mathrm{abs}}$, 2D $\mathcal{P}_{\mathrm{RoPE}}$ & \textbf{74.88} & \textbf{58.7} & \textbf{61.9} & \textbf{63.3} & \textbf{82.9} \\
1D $\mathcal{P}_{\mathrm{abs}}$, 2D $\mathcal{P}_{\mathrm{RoPE}}$ & 74.24 & 57.8 & 61.7 & 61.2 & 80.8 \\
1D $\mathcal{P}_{\mathrm{abs}}$, 1D $\mathcal{P}_{\mathrm{RoPE}}$ & 74.12 & 57.2 & 60.8 & 59.8 & 79.8 \\
\bottomrule
\end{tabular}
\label{tab:pos}
\end{table}

\subsection{2D Absolute Positional Encoding with Shared Coordinate System}

Unlike traditional approaches that flatten image patches into 1D sequences and use learnable 1D embeddings, we explicitly maintain the 2D grid structure and enforce a shared canonical token-grid coordinate system between the template and the search region.

\begin{table*}[t]
\small
\renewcommand{\arraystretch}{1.18}
\setlength\tabcolsep{5.2pt}
\centering
\caption{Comparison with state-of-the-art trackers under BBOX initialization on TNL2K, LaSOT, and OTB99. The best and second-best results are highlighted in bold and underlined, respectively.}
\begin{tabular}{c|c|ccc|ccc|ccc}
\toprule
\multirow{2}{*}{Tracker} & \multirow{2}{*}{Initialize} & \multicolumn{3}{c|}{TNL2K} & \multicolumn{3}{c|}{LaSOT} & \multicolumn{3}{c}{OTB99} \\ \cline{3-11}
& & AUC & Prec & NPrec & AUC & Prec & NPrec & AUC & Prec & NPrec \\ \hline
SiamRPN++~\cite{ref37} & BBOX & 41.3 & 41.2 & 48.0 & 49.6 & 49.1 & 56.9 & - & - & - \\
AutoMatch~\cite{ref38} & BBOX & 47.2 & 43.5 & - & 58.3 & 59.9 & 67.4 & - & - & - \\
TrDiMP~\cite{ref39} & BBOX & 52.3 & 52.8 & - & 63.9 & 61.4 & - & - & - & - \\
TransT~\cite{ref40} & BBOX & 50.7 & 51.7 & - & 64.9 & 69.0 & 73.8 & - & - & - \\
SwinTrack-B~\cite{ref41} & BBOX & - & 57.1 & - & \underline{69.6} & 74.1 & 78.6 & - & - & - \\
OSTrack-256~\cite{ref42} & BBOX & 54.3 & - & - & 69.1 & \underline{75.2} & 78.7 & - & - & - \\
UVLTrack-B~\cite{ref2} & BBOX & 62.7 & 65.4 & - & 69.4 & 74.9 & - & 69.3 & 90.1 & 84.3 \\
MambaVLT~\cite{ref27} & BBOX & \textbf{63.3} & \underline{65.8} & \textbf{87.5} & 65.0 & 69.5 & 76.6 & \textbf{71.6} & 92.9 & 87.4 \\
SAVLT-B~\cite{ref34} & BBOX & \textbf{63.3} & \textbf{67.3} & 80.3 & \textbf{70.6} & \textbf{76.4} & \textbf{82.1} & 70.2 & 90.8 & - \\
\rowcolor{ourscolor} \textbf{LVTrack-256} & BBOX & 61.5 & 64.0 & 79.7 & 66.2 & 69.8 & 78.1 & 70.1 & \underline{93.8} & \underline{87.6} \\
\rowcolor{ourscolor} \textbf{LVTrack-384} & BBOX & \underline{62.9} & 66.3 & \underline{80.4} & 68.0 & 71.3 & \underline{78.9} & \underline{71.0} & \textbf{93.9} & \textbf{87.8} \\
\bottomrule
\end{tabular}
\label{tab:ptrack}
\end{table*}
\textbf{Formulation.} Let $\mathcal{P}_{\mathrm{abs}} \in \mathbb{R}^{H_{\mathrm{grid}} \times W_{\mathrm{grid}} \times D}$ be a learnable parameter grid, where $H_{\mathrm{grid}}$ and $W_{\mathrm{grid}}$ correspond to the feature map dimensions of the search region ($16 \times 16$ tokens in our model), and $D$ is the feature dimension.
For the search region, the spatial dimensions typically align with the grid size. Thus, the positional encoding $P^s$ is directly indexed from the grid:

\begin{equation}
    P^s_{(i,j)} = \mathcal{P}_{\mathrm{abs}}[i, j], \quad \text{for } 0 \le i < H_{\mathrm{grid}}, 0 \le j < W_{\mathrm{grid}}.
\end{equation}

For the template, since its resolution is typically smaller than the search region ($8 \times 8$ tokens in our model), we do not learn a separate set of embeddings. Instead, to ensure geometric continuity and consistency, we derive the template position codes via bilinear interpolation from $\mathcal{P}_{\mathrm{abs}}$. Let $\Phi(\cdot)$ denote the interpolation function:

\begin{equation}
    P^t = \Phi(\mathcal{P}_{abs}, \text{size}=(H_t, W_t)),
\end{equation}
where $(H_t, W_t)$ are the dimensions of the template feature map. This implies that the positional distribution of the template is treated as a resampled subset of the search region's coordinate system.

\subsection{2D Axial Rotary Positional Embedding}

To enhance the model's perception of relative positions—crucial for handling object translation and deformation—we apply 2D RoPE to the query and key vectors in the attention layers \cite{ref49}.

\textbf{Formulation.} Standard RoPE is defined for 1D sequences. To adapt it for 2D images, we split the feature channels $D$ into two halves to encode the $y$-axis (height) and $x$-axis (width) information, respectively. For a feature vector $\mathbf{x} \in \mathbb{R}^D$ at spatial index $(h, w)$, we decompose it as $\mathbf{x} = [\mathbf{x}^{(h)} || \mathbf{x}^{(w)}]$, where $\mathbf{x}^{(h)}, \mathbf{x}^{(w)} \in \mathbb{R}^{D/2}$.
We apply the standard rotary transformation to each half. Taking the $x$-axis component as an example, for the $k$-th pair of elements, the rotation angle is $\theta_k \cdot w$, with $\theta_k = 10000^{-2k/(D/2)}$. The transformation is defined as:

\begin{equation}
    \text{RoPE}_{\mathrm{1D}}(\mathbf{x}^{(w)}, w) = \begin{pmatrix} \cos(w\theta_k) & -\sin(w\theta_k) \\ \sin(w\theta_k) & \cos(w\theta_k) \end{pmatrix} \begin{pmatrix} x^{(w)}_{2k} \\ x^{(w)}_{2k+1} \end{pmatrix}.
\end{equation}
The final 2D relative position injection is the concatenation of the transformed parts:

\begin{equation}
    \tilde{\mathbf{q}}_{(h,w)} = \text{Concat}(\text{RoPE}_{\mathrm{1D}}(\mathbf{q}^{(h)}, h), \text{RoPE}_{\mathrm{1D}}(\mathbf{q}^{(w)}, w)).
\end{equation}
This operation is applied independently to the tokens of the template and the search region, preserving their respective intra-frame local topology.

As shown in Tab. \ref{tab:pos}, our ``2D $\mathcal{P}_{\mathrm{abs}}$ + 2D $\mathcal{P}_{\mathrm{RoPE}}$" scheme achieves the best performance.
The experimental results show that 1D $\mathcal{P}_{\mathrm{RoPE}}$ performs significantly worse than 2D $\mathcal{P}_{\mathrm{RoPE}}$.
This is because 1D encoding destroys the 2D neighborhood structure of images.
In a 1D sequence, index $k$ is adjacent to $k+1$, but far from $k+W$ (vertical neighbor).
2D RoPE makes the attention score explicitly dependent on relative offsets along both spatial axes, which better preserves the 2D neighborhood structure than 1D sequence encoding.
This property is vital for tracking, as the object translates within the search region.
2D RoPE ensures that the relative geometric structure between feature points remains translation-consistent regardless of the object's absolute position, thereby enhancing robustness.

In summary, our hybrid strategy leverages $\mathcal{P}_{\mathrm{abs}}$ to provide a globally unified coordinate reference and $\mathcal{P}_{\mathrm{RoPE}}$ to preserve translation-invariant local geometry, facilitating effective feature interaction between the template and the search region.

\section{BBOX-initialized Tracking Performance}
\label{app:pure_tracking}

Although LVTrack is mainly designed for NL-initialized and NL+BBOX-initialized settings, it also supports the conventional BBOX-initialized protocol.
This flexibility stems from the language-injection architecture, which allows linguistic conditioning to be cleanly enabled or omitted without altering the core tracking pipeline.
In the BBOX-initialized setting ($m=\mathcal{S}$), the first-frame ground-truth box is used to initialize tracking, and the injector is bypassed so that the model runs without relying on linguistic cues.
Tab. \ref{tab:ptrack} compares LVTrack with representative box-initialized trackers and three recent vision-language trackers (UVLTrack-B, MambaVLT, and SAVLT-B) under the same initialization.

Overall, LVTrack achieves competitive performance across TNL2K, LaSOT, and OTB99.
LVTrack-384 obtains the second-best AUC and NPrec on TNL-2K (62.9/80.4), the second-best NPrec on LaSOT (78.9), and the second-best AUC on OTB99 (71.0), while achieving the best Prec and NPrec on OTB99 (93.9/87.8).
LVTrack-256 also performs strongly, ranking second on OTB99 Prec and NPrec (93.8/87.6).
Although SAVLT-B leads on LaSOT under the pure BBOX protocol, LVTrack remains competitive with recent vision-language trackers and surpasses most earlier siamese/transformer baselines.
These results confirm that LVTrack is not restricted to language-driven initialization and provides good support for box-initialized tracking as well.

\section{More Visualizations}
\begin{figure*}[t]
\centering
\includegraphics[width=6.2in]{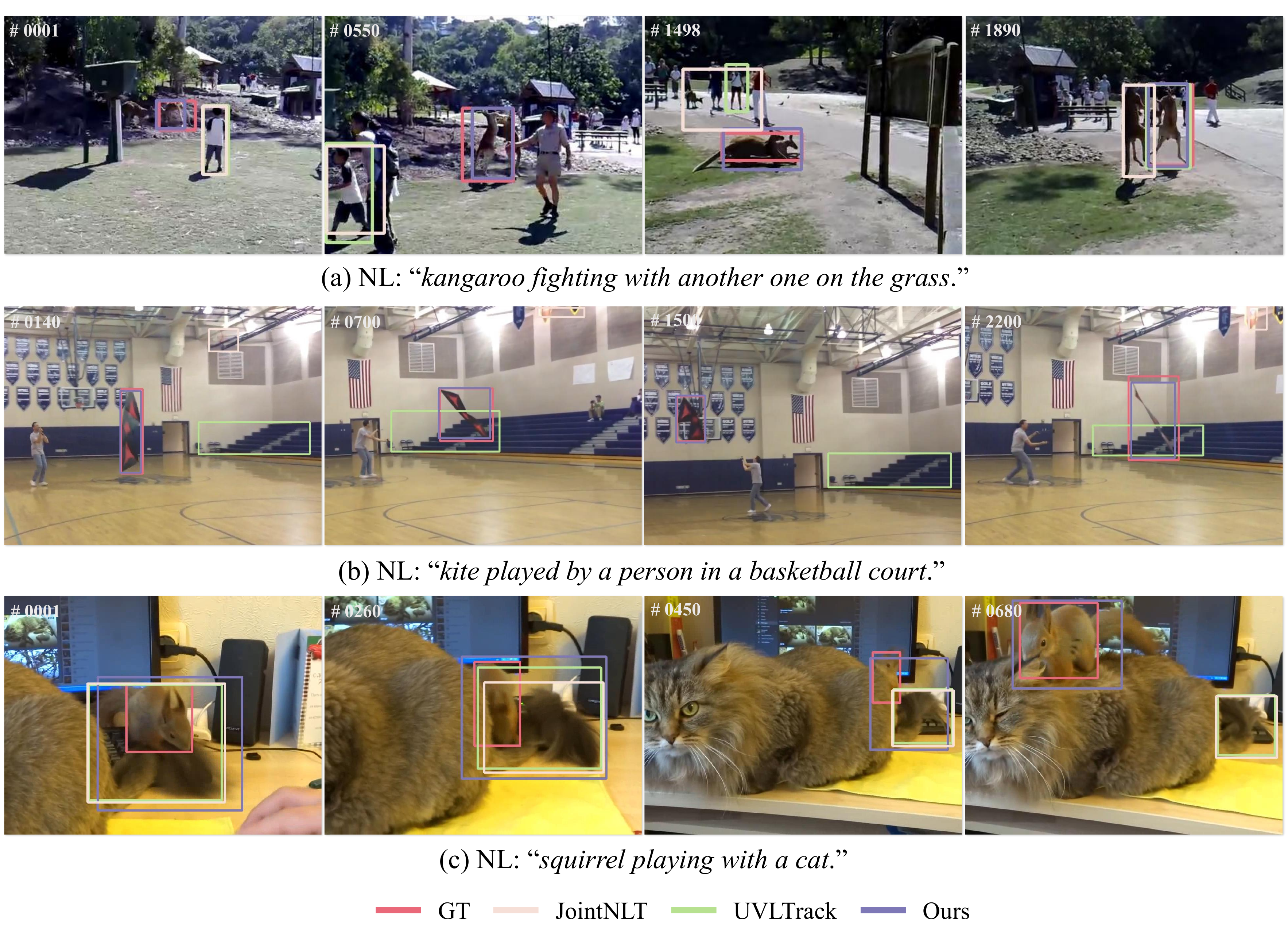}
\hfil
\caption{Additional Visualizations under NL Initialization.}
\label{vis_add}
\end{figure*}

We present more tracking visualizations and comparative results in Fig.~\ref{vis_add}, which demonstrate the tracking performance under challenging conditions such as occlusion and appearance variations.
In Fig.~\ref{vis_add}(a), LVTrack accurately localizes the target ``kangaroo'', whereas the competing trackers fail to perform correct language-guided grounding and instead drift to nearby persons. 
In Fig.~\ref{vis_add}(b), LVTrack remains robust to a flying kite whose appearance and shape vary continuously over time, demonstrating its strong high-level semantic understanding and adaptation capability. 
Fig.~\ref{vis_add}(c) further illustrates a challenging scenario with severe distractors and occlusion, where the target is almost completely lost. 
When the ``squirrel'' reappears, LVTrack successfully re-localizes and locks onto the target, showing its effective use of language cues and strong robustness against interference.
